\documentclass[letterpaper, 10 pt, conference]{ieeeconf} 
\IEEEoverridecommandlockouts

\usepackage{amsmath,amsfonts}
\usepackage{array}
\usepackage{cite}
\usepackage[caption=false,font=scriptsize]{subfig}
\usepackage{textcomp}
\usepackage{stfloats}
\usepackage{url}
\usepackage{verbatim}
\usepackage{graphicx}
\usepackage{amssymb}
\usepackage{pifont}
\usepackage[]{todonotes} % disabled because it breaks tikzexternalize
\usepackage{hyperref}
\usepackage[capitalize]{cleveref}
\usepackage{comment}
\usepackage{multirow}
\usepackage{booktabs}
\usepackage{rotating}
\usepackage{makecell}
\usepackage{tabularx}
\usepackage{xcolor}
\usepackage[linesnumbered]{algorithm2e}
\RestyleAlgo{ruled}
\DontPrintSemicolon 

\usepackage{blindtext}
\usepackage{siunitx}
\AtBeginDocument{\RenewCommandCopy\qty\SI}
\usepackage{tikz}
\usetikzlibrary{arrows.meta}
\usetikzlibrary{arrows}
\usepackage{pgfplots}
\pgfplotsset{compat=1.18}
\usepackage{xcolor}
\definecolor{Bluelight}{HTML}{0065BD} % TUMBlue
\definecolor{Black}{HTML}{000000}
\definecolor{Blue}{HTML}{005293}
\definecolor{Bluestrong}{HTML}{003359}
\definecolor{Red}{HTML}{8C000F}
\definecolor{Grey}{HTML}{808080}
\definecolor{Greylight}{HTML}{CCCCCC}
\definecolor{Orange}{HTML}{E37222}
\definecolor{Green}{HTML}{A2AD00}
\definecolor{GreenCR}{HTML}{008000}
\definecolor{OrangeCR}{HTML}{f1b514}
\definecolor{visibleArea}{HTML}{81AF82}
\definecolor{occludedAreaObst}{HTML}{DC8282}
\definecolor{occludedAreaLanelet}{HTML}{DBBC82}

\def\BibTeX{{\rm B\kern-.05em{\sc i\kern-.025em b}\kern-.08em
    T\kern-.1667em\lower.7ex\hbox{E}\kern-.125emX}}

\begin{document}

%\title{\LARGE \bf Unsupervised Learning for Detection of Rare Driving Scenarios}

%\author{Dat Le$^{1}$, Thomas Manhardt$^{2}$, Moritz Venator$^{2}$, Johannes Betz$^{3}$

%\thanks{$^{3}$ Johannes Betz is with the Professorship of Autonomous Vehicle Systems, TUM School of Engineering and Design, Technical University of Munich, 85748 Garching, Germany; Munich Institute of Robotics and Machine Intelligence (MIRMI). Corresponding author: johannes.betz@tum.de}

%\thanks{$^{2}$ Thomas Manhardt and Moritz Venator are with CARIAD SE, In-Campus Allee 22, 85053 Ingolstadt, Germany.}

%\thanks{$^{1}$ Dat Le is with the TUM School of Computation, Information and Technology, Technical University of Munich, 85748 Garching, Germany.}}

\title{\LARGE \bf Unsupervised Learning for Detection of Rare Driving Scenarios}
\author{Dat Le$^{1}$, Thomas Manhardt$^{2}$, Moritz Venator$^{2}$, Johannes Betz$^{1}$  % <-this % stops a space
%\thanks{Manuscript received XXX, 2022; revised XXX 2022. \textit{(Corresponding author: Tobias Betz (email: tobi.betz@tum.de)}}
\thanks{$^{1}$ D. Le and J. Betz are with the Professorship of Autonomous Vehicle Systems, TUM School of Engineering and Design, Technical University Munich, 85748 Garching, Germany; Munich Institute of Robotics and Machine Intelligence (MIRMI), \{{johannes.betz}\}@tum.de}
\thanks{$^{2}$ T. Manhardt and M. Venator are with CARIAD SE, In-Campus Allee 22, 85053 Ingolstadt, Germany.}}% <-this % stops a space



\maketitle
\thispagestyle{empty} %--to make title page number less
\pagestyle{empty}    % -- to make other pages number less.
\begin{abstract}

The detection of rare and hazardous driving scenarios is a critical challenge for ensuring the safety and reliability of autonomous systems. This research explores an unsupervised learning framework for detecting rare and extreme driving scenarios using naturalistic
driving data (NDD). We leverage the recently proposed  Deep Isolation Forest (DIF), an anomaly detection algorithm that combines neural network-based feature representations with Isolation Forests (IFs), to identify non-linear and complex anomalies. Data from perception modules, capturing vehicle dynamics and environmental conditions, is preprocessed into structured statistical features extracted from sliding windows. The framework incorporates t-distributed stochastic neighbor embedding (t-SNE) for dimensionality reduction and visualization, enabling better interpretability of detected anomalies. Evaluation is conducted using a proxy ground truth, combining quantitative metrics with qualitative video frame inspection. Our results demonstrate that the proposed approach effectively identifies rare and hazardous driving scenarios, providing a scalable solution for anomaly detection in autonomous driving systems. Given the study's methodology, it was unavoidable to depend on proxy ground truth and manually defined feature combinations, which do not encompass the full range of real-world driving anomalies or their nuanced contextual dependencies.

\end{abstract}

%\vspace{0.8em}
%\begin{IEEEkeywords}
%Driving Anomaly Detection, Deep Isolation Forest, ADAS, t-SNE.
%\end{IEEEkeywords}
%
%!TeX spellcheck = en_US
% !TeX root = ../main.tex

\section{Introduction}
\label{sec:introduction}

\begin{figure}
    \centering
    \includegraphics[width=1\linewidth]{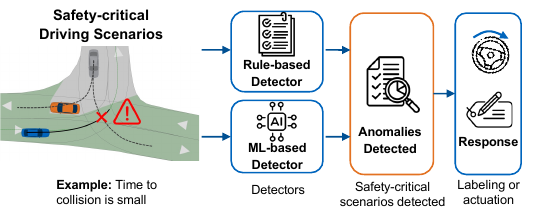}
    \caption{Illustration of the pipeline for detecting safety-critical driving scenarios, showcasing the integration of ML-based and rule-based detectors to identify anomalies and respond effectively.}
    \label{fig:concept_idea}
\end{figure}

Advanced Driver Assistance Systems (ADAS) have greatly improved road safety and driver comfort by handling challenging situations and avoiding potential hazards. A significant challenge for these systems is determining when to intervene \cite{wang2024esp}. Although certain situations, such as lane departure, clearly require warning or feedback, it can be challenging to identify scenarios that deviate from normal driving conditions (Fig.~\ref{fig:concept_idea}).

Traditionally, detecting driving anomalies has relied heavily on expert domain knowledge and manual analysis of driving data. Numerous rule-based approaches have been proposed for driving anomaly detection, including abnormal driving behaviors \cite{5482295,6232298,7338354}, risky driving scenarios \cite{7116534,7795864}, and monitoring road conditions \cite{inproceedings1,9095790,article1}. This process is labor-intensive, time-consuming, and challenging to scale as data volumes grow. Exhaustively listing all hazardous scenarios is impractical, and manual analysis often misses subtle or complex patterns, limiting anomaly detection.  In contrast, data-driven methods using machine learning excel at detecting non-linear and complex anomalies and can effectively handle diverse data features, making them suitable for dynamic driving environments. However, these models often require extensive labeled datasets for training and evaluation, which is impractical in real-world scenarios due to the scale of the dataset and the rarity of certain anomalies.

%It is nearly impossible to exhaustively tabulate all possible actions or situations that lead to hazardous scenarios. Moreover, manual analysis frequently overlooks subtle or complex patterns, limiting its ability to capture potential anomalies. 

To address these limitations, we propose an automated framework leveraging unsupervised learning to detect driving anomalies without relying on labeled data. These anomalies include a variety of types, ranging from point anomalies (e.g., sensor errors) to contextual anomalies and collective patterns in driving scenarios \cite{9304789}. Our study primarily focuses on rare, extreme, and abnormal events, such as sensor malfunctions and risky maneuvers. Unlike traditional rule-based approaches, which rely on pre-defined conditions \cite{7338354,inproceedings2}, our method identifies non-intuitive patterns that cannot easily be categorized. A key challenge in this work is evaluating unsupervised learning models when labeled data is unavailable. As mentioned earlier, creating labeled data is both time-consuming and impractical, especially for extreme driving scenarios. To address this, we propose the use of a proxy ground truth set consisting of approximate annotations derived from domain knowledge. This proxy set allows us to evaluate our unsupervised learning framework effectively without labeled data. Additionally, we utilize t-distributed Stochastic Neighbor Embedding (t-SNE) \cite{JMLR:v9:vandermaaten08a}, a dimensionality reduction technique, to visualize high-dimensional data in 2D, making it easier to interpret detected anomalies. The main contributions of our study are:

\begin{itemize}
    \item  We apply the recently proposed Deep Isolation Forest (DIF) in a novel unsupervised framework for detecting anomalies in street scenarios. 
    
    \item We implement a robust data preprocessing pipeline tailored to naturalistic driving data., including selecting and engineering features from multimodal driving data (e.g., vehicle bus signals, object detection, lane detection). 
    
    \item We provide a robust evaluation and visualization of our anomaly detection framework by providing driving anomalies using proxy ground truth and t-SNE.
\end{itemize}

%!TeX spellcheck = en_US
% !TeX root = ../main.tex

\section{Related Work}
\label{sec:relatedwork}

\subsection{Driving Anomaly Detection}
\label{subsec:drivinganomalydetection}

Anomaly detection is the process of identifying unexpected events that deviate from the norm. The types of anomalies range from point anomalies and collective anomalies to more complex contextual anomalies. In automated driving, Breitenstein et al. \cite{9304789} provided a systematization of anomalies for visual perception, with the categories structured by detection complexity. Heidecker et al. \cite{DBLP:journals/corr/abs-2103-03678} even extend the categorization of anomalies for perception concerning camera, LiDAR, and RADAR sensor modalities. 

Studies have proposed anomaly detection approaches in particular problems by setting thresholds. Malta et al. \cite{4815488} proposed an anomaly detection method based on vehicle speed and brake pedal thresholds. The scenarios are considered dangerous when the mean velocity is above a certain threshold, and the brake pedal pressure is high. Zhao et al. \cite{6831046} used acceleration and steering wheel information to identify aggressive driving behavior. They set the thresholds based on the angles of the steering wheel. These established rule-based models are only effective for simple cases. 
When the driving environment is complicated, they may underestimate risk when the thresholds are not met or overstate risk even when drivers control the car properly.

An alternative approach is to detect anomalies using machine learning algorithms. Hofmockel et al. \cite{inproceedings3} and Matousek et al. \cite{8417777} implement Isolation Forest to detect anomalous driving behaviors such as aggressive maneuvers, drowsiness, and tailgating, using raw vehicle sensor data via the Controller Area Network (CAN-bus). The model shows its effectiveness across different benchmarks and strong scalability but fails to detect hard anomalies that are difficult to isolate in non-linear separable data space. Chen et al. \cite{7338354} introduced an SVM-based method to classify the types of abnormal driving behaviors. The study utilized orientation data from a smartphone. However, due to the complexity and variability of driving anomalies, creating a comprehensive classification-based solution that accurately identifies all anomalous scenarios is challenging and time-intensive. Clustering-based methods have also been employed for driving anomaly detection. Zheng et al. \cite{7795771} proposed an unsupervised clustering technique utilizing smartphone accelerometer sensor data. The outliers on the clustering map were considered anomalies. Still, identifying meaningful clusters becomes problematic as the dimensionality of the feature space grows.

\subsection{Isolation Forest and Its Extensions}
\label{subsec:isolationforestanditsextensions}
Isolation Forest (IF) \cite{4781136} is widely used for anomaly detection due to its simplicity, efficiency, and scalability. It performs well across various benchmarks and is particularly suited for large datasets \cite{bouman2023unsupervisedanomalydetectionalgorithms}. 

The core assumption of IF is that anomalies, being "few and different", are easier to isolate from the rest of the data than normal instances. The algorithm constructs an ensemble of binary decision trees, known as isolation trees, built by recursively applying random splits on the data until all instances are isolated. Anomalies typically result in shorter average path lengths within these trees. One major limitation of IF is that it employs a linear, axis-parallel isolation approach, where each split considers only one feature at a time, making it hard to effectively detect complex anomalies in non-linear separable data spaces. 

\begin{figure}[htbp]
    \centering
    \includegraphics[width=0.95\linewidth]{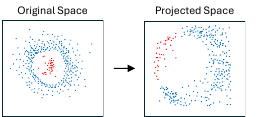}
    \caption{Synthetic data example showing hard anomalies in the original space (left) and their representation in the projected space by DIF (right).}
    \label{fig:hard_anomalies}
\end{figure}

Several extensions of IF have been proposed to address this. SCIF \cite{Liu2010OnDC} introduces a non-axis-parallel branching criterion by employing optimal slicing hyperplanes rather than single-feature splits. EIF \cite{Hariri_2021} also uses hyperplanes but with random slopes and intercepts, enabling more diverse partitions. To mitigate the empty branching issue in EIF, Lesouple et al. \cite{article5} select splitting thresholds from the range of projected values along the hyperplane direction. A probability-based method from Tokovarov et al. \cite{10.1016/j.ins.2021.10.075} also refines split selection by finding more effective splitting values than random choices. These extensions generally improve performance by using non-axis-parallel or heuristic partitions. However, the main issue is that they still rely on shallow, linear isolation techniques. While these methods can approximate non-linear partitions through recursive splits, isolating hard anomalies (as shown in Fig. \ref{fig:hard_anomalies}) often requires many splits, resulting in long path lengths, which means lower anomaly scores and are, therefore, not considered anomalies. This prevents true anomalies from being detected, leading to high false negative errors.

Deep Isolation Forest (DIF) \cite{10108034} addresses this limitation by introducing a novel approach, which integrates non-optimized deep neural networks (DNNs) with Isolation Forests (IFs). The randomly initialized DNNs project the original data into random representation spaces. (an example shown in Fig. \ref{fig:hard_anomalies}). Then, IFs use simple axis-parallel cuts to identify anomalies in these new data spaces. The main goal of the randomly initialized DNNs is to better detect complex anomalies that are not easily isolated in the original non-linear data space.

%By liberating the algorithm from linear constraints, DIF significantly improves its capacity to detect subtle, complex anomalies that are otherwise hidden in the original data space while maintaining the scalability and efficiency of the IF’s framework.

%!TeX spellcheck = en_US
% !TeX root = ../main.tex
\newpage 
\section{Methodology}
\label{sec:method}
%This section is structured as follows. Section III-A describes the Data Processing and Feature Engineering. The Deep Isolation Forest, a novel method for anomaly detection, is introduced in section III-B. And section III-C explained t-SNE, dimensionality reduction.
\subsection{Anomaly Detection Framework}
\label{subsec:workflow}
\begin{figure*}[htbp]
    \centering
    \includegraphics[width=\textwidth]{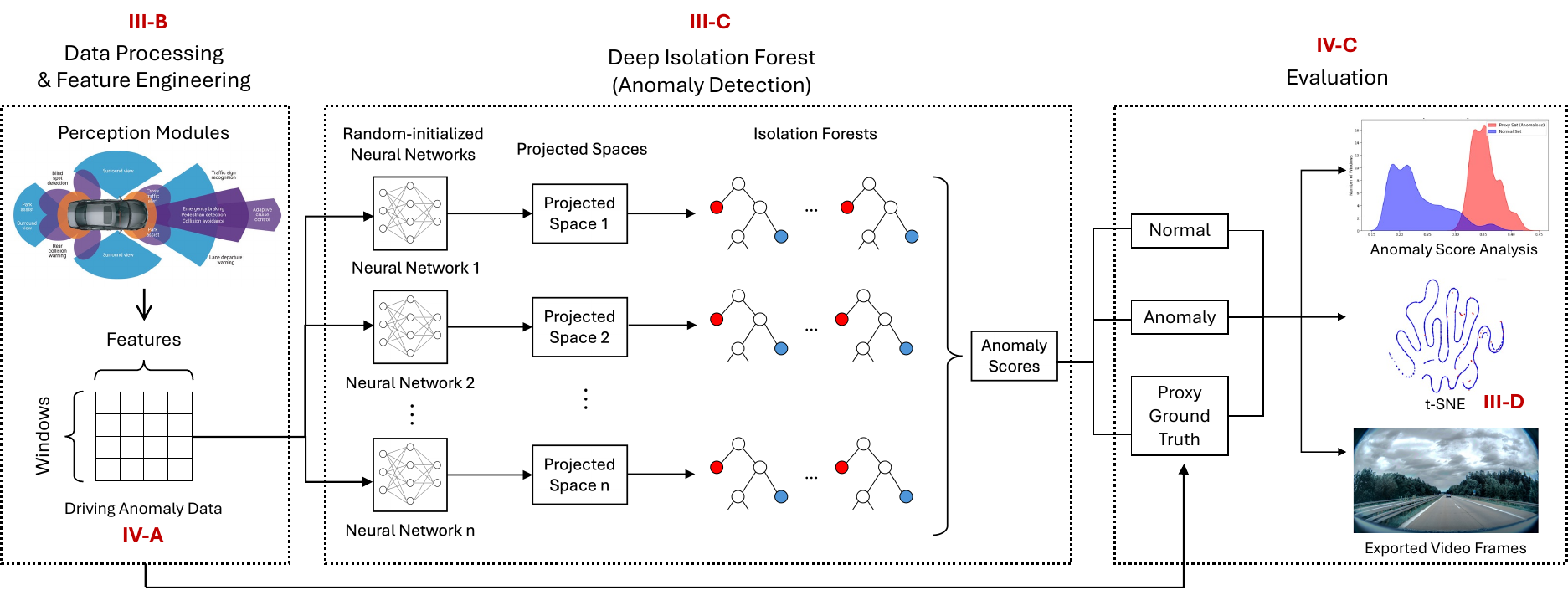} % Replace with the path to your image
    \caption{Overview of the novel anomaly detection framework: The figure depicts the flow of \textit{vehicle bus signals} and \textit{perception signals} (derived from \textit{perception modules}) through processing and feature engineering steps, followed by anomaly detection using Deep Isolation Forest. The model takes Driving Anomaly Data as input, a multivariate tabular dataset where rows represent windows and columns represent features. The output consists of anomaly scores for each window. A threshold is defined to classify windows as anomalies if their anomaly scores exceed the threshold, while those below it are considered normal. Finally, the detection results are evaluated using proxy ground truth set derived from the Driving Anomaly Data.}
    \label{fig:overview_workflow} % Replace with your desired label
\end{figure*}

The overall workflow of our framework is illustrated in Fig.~\ref{fig:overview_workflow}. We utilize naturalistic driving data (NDD), consisting of \textit{vehicle bus signals} and \textit{perception outputs}. Vehicle bus signals are sensor data capturing internal vehicle dynamics, such as speed, acceleration, and yaw rate. Perception outputs, derived from environmental perception modules like object detection, blindness detection, and lane detection, provide contextual information about the surrounding environment. These data, stored as a multivariate time series, is processed and feature-engineered into a multivariate tabular format, referred to as Driving Anomaly Data (DAD), and used as input for unsupervised anomaly detection. Our study primarily employs Deep Isolation Forest (DIF) \cite{10108034}, an advanced anomaly detection algorithm that integrates neural networks, and Isolation Forest (IF) \cite{4781136} to capture complex feature interactions and detect non-linear patterns. For unsupervised evaluation, we created a rule-based proxy dataset, considered as ground-truth anomalies, and employed a three-pronged evaluation strategy: i) analyzing the anomaly score distributions of the proxy and normal sets, comparing the top 100 highest-score anomaly segments with randomly selected normal segments, ii) visualizing data and reducing dimensionality using t-SNE \cite{JMLR:v9:vandermaaten08a}, and iii) exporting video frames of detected anomalies for perceptual evaluation.

\subsection{Data Processing and Feature Engineering}
\label{subsec:dataprocessingandfeatureengineering}

\begin{figure*}[htbp]
    \centering
    \includegraphics[width=\textwidth]{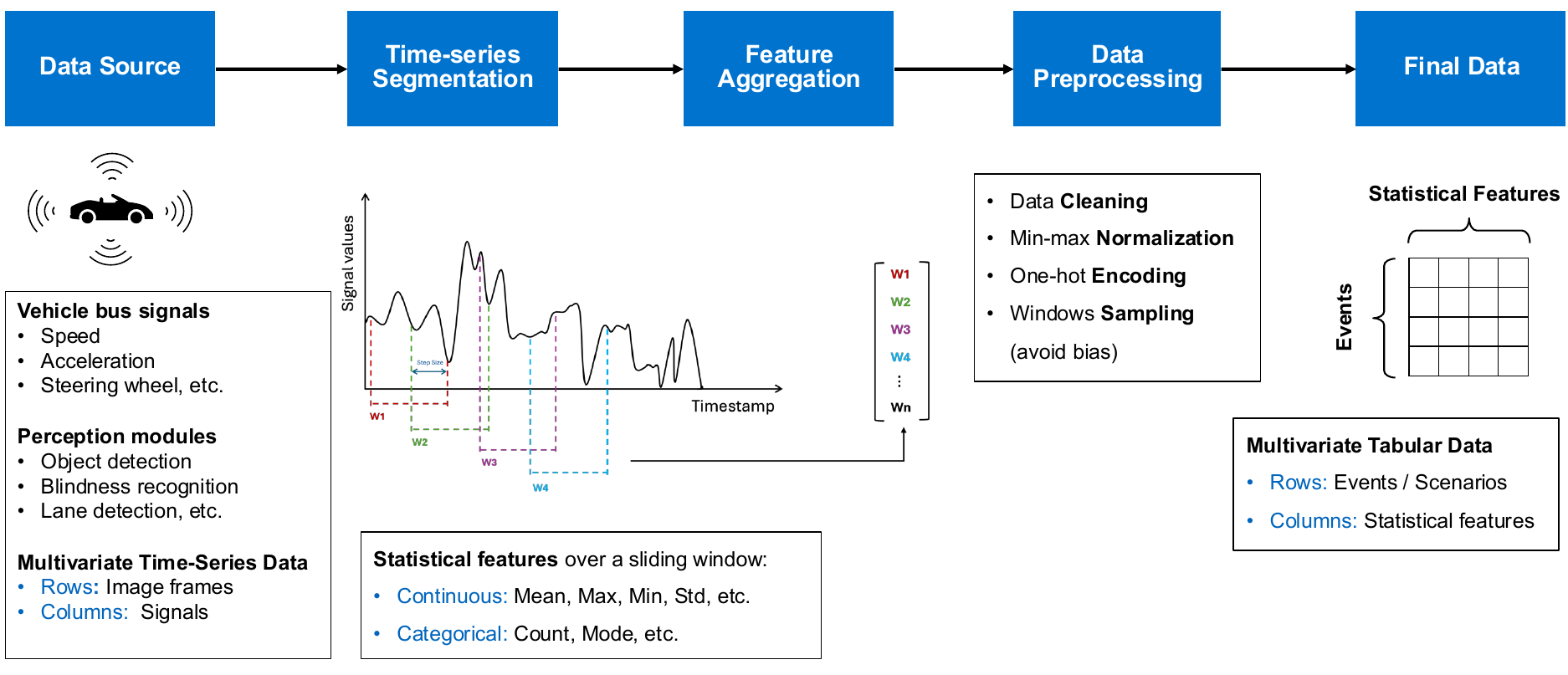} % Replace with the path to your image
    \caption{Data processing and feature engineering pipeline: Multivariate time-series signals from vehicle and perception modules are segmented into windows, aggregated into statistical features, and preprocessed into a tabular dataset for anomaly detection.}
    \label{fig:overview_data_Processing} 
\end{figure*}

To enable effective anomaly detection, raw multimodal signals must be transformed into a structured, model-ready format. Our data processing pipeline (Fig. \ref{fig:overview_data_Processing}) begins with time-series data collected from both vehicle bus signals and perception modules.We utilized 100 hours of naturalistic driving data (NDD) collected from multiple measurements recorded during test drives on public roads in Europe. The dataset is multimodal, real-world, and extracted from perception systems, comprising modules such as CAN bus, blindness detection, object detection, and lane detection. The data is structured as a multivariate time series, where rows represent frames (sampled at approximately 10 Hz), and columns correspond to perception module signals. The first step was to identify and select relevant signals that capture vehicle kinematics and environmental factors. For vehicle kinematics, we selected speed as a fundamental and essential signal. For environmental factors, we included road-type conditions, weather, and time-to-collision (TTC) with detected vehicles. 

%Further discussion on signal selection and features is provided in Section~\ref{sec:dataset}.

We employed a feature aggregation approach combining sliding window segmentation and feature extraction. Sliding window segmentation divides the time series data into overlapping windows of fixed duration, allowing for the representation of localized temporal behavior. In this study, we used a window size of 6 seconds with a step size of 3 seconds, resulting in a 50\% overlap between consecutive windows. Within each window, we extract features to summarize the behavior of the signals. The extracted features, described in Table~\ref{tab:features_characteristics}, range from simple statistical measures to complex derived metrics that quantify higher-order driving behaviors and risks. Continuous features were standardized (mean of zero and unit variance) to maintain comparable scales across the dataset. Categorical features were encoded into binary indicators using one-hot encoding, ensuring compatibility with the anomaly detection models. In order to mitigate biases caused by temporal continuity, the windows were randomly sampled across different time series measurements. This ensures that the final multivariate tabular dataset represents a diverse range of driving scenarios, minimizing dependence on sequential patterns and improving model generalizability.

\subsection{Deep Isolation Forest}
\label{subsec:deepisolationforestmethod}

Deep Isolation Forest (DIF) \cite{10108034} is a recently introduced hybrid algorithm  that combines random neural network-based representations with axis-parallel isolation trees to enhance anomaly detection by creating non-linear partitions in transformed data spaces. The architecture of DIF is illustrated in Fig.~\ref{fig:overview_workflow}. First, the random initialized (i.e., non-optimized) neural networks project the original data into multiple random representation spaces. This random representation ensemble is defined as:
%\begin{equation}
%    \Phi = \{\phi_u : D \to \mathbb{R}^d \mid u = 1, 2, \ldots, n\},
%\end{equation}
\begin{equation}
\Omega(D) = \{\chi_u \subset \mathbb{R}^d \mid \chi_u = \phi_u(D; \theta_u)\}_{u=1}^n
\label{eq:representation_function}
\end{equation}
where \( \chi_u \) is the \( u \)-th random representation, \( \phi_u \) is the \( u \)-th neural network that maps the original input data \( D \) into a new \( d \)-dimensional space (with \( d \) controlling the richness of the representation and chosen based on the task or model design), \( n \) is the ensemble size (number of neural networks), and \( \theta_u \) are the randomly initialized network weights.

Each representation \( \chi_u \) is assigned with \(t\) isolation trees (iTrees), and a forest \(\Gamma = \{\tau_i\}_{i=1}^T\) containing \(T = n \times t\) iTrees is constructed. An iTree \(\tau_i\) is essentially a binary tree. An ensemble of iTrees constitutes an Isolation Forest (IF), which iteratively partitions the data using random splits to isolate anomalies. An iTree consists of a subset of data starting at the root node and proceeding to an isolated leaf node by recursively splitting the data into smaller subsets until each sample is isolated. This concept assumes that anomalies require fewer splits to be separated from other normal samples. IF then assigns an anomaly score by calculating the average path length needed to isolate a sample across all iTrees. For each sample, the path length is defined as the number of edges traversed from the root node to the isolated leaf node. Since abnormal samples are expected to require fewer splits to be isolated, they tend to have a shorter average path. The anomaly score for a given sample is calculated as follows:
\begin{equation}
    s(x,m) = 2^{-\frac{E(h(x))}{c(m)}},
\end{equation}
where \(s(x,m)\) is the anomaly score for sample \(x\), \(h(x)\) is the path length of \(x\) in a single iTree, \(E(h(x))\) is the average path length of \(x\) across all iTrees, and \(c(m)\) is the normalization factor for a dataset of size \(m\). The normalization factor \(c(m)\) is given by:
\begin{equation}
    c(n) = 2H(m-1) - \frac{2(m-1)}{m},
\end{equation}
where \(H(i)\) is the \(i\)-th harmonic number, approximated as:
\begin{equation}
    H(i) \approx \ln(i) + 0.5772156649,
\end{equation}
with \(0.5772156649\) being the Euler-Mascheroni constant. The anomaly score \(s(x,m)\) ranges from 0 to 1, where values closer to 1 indicate higher anomaly likelihood.

\subsection{t-SNE for Visualization}
\label{subsec:t-sne}
We use t-distributed Stochastic Neighbor Embedding (t-SNE) \cite{JMLR:v9:vandermaaten08a} for qualitative visualization of anomaly detection results in 2D.  This is an unsupervised non-linear dimensionality reduction technique that visualizes high-dimensional data by mapping it to a low-dimensional space, typically 2D or 3D. Unlike Principal Component Analysis (PCA) \cite{doi:10.1080/14786440109462720}, which is a linear dimensionality reduction technique suited for data with a linear structure, t-SNE is a non-linear method designed to preserve pairwise similarities between data points in a lower-dimensional space. In our framework, we apply t-SNE to the aggregated feature data to qualitatively inspect how anomalous and normal windows are separated. This helps assess how well the model distinguishes rare scenarios. We use a standard implementation with default settings and do not modify the algorithm. %While PCA aims to maximize variance by preserving large pairwise distances, t-SNE focuses on maintaining small pairwise distances to better capture the local structure of the data. t-SNE finds similarity measures between pairs of instances in both high-dimensional and low-dimensional spaces, then optimizes these similarities as follows: 

\section{Experiments \& Results}
\label{sec:results}

\subsection{Dataset}
\label{sec:dataset}

\begin{table*}[h]
\renewcommand{\arraystretch}{2}
\caption{Key features of signal characteristics.}
\label{tab:features_characteristics}
\centering
\begin{tabular}{|c|p{5cm}|p{5cm}|p{4cm}|}
\hline
\textbf{Aspects}       & \textbf{Features}                                   & \textbf{Values}                                  & \textbf{Type}                       \\ \hline
Vehicle Kinematics       & Relative Speed Range                               & \(\frac{max(speed) - min(speed)}{max(speed)}\) \vspace{0.2cm}                        & Continuous                          \\ \hline
Environmental factor      & Rain severe \par Sunray severe \par Camera image blurriness severe \par Road-type conditions
                         & Severe, Normal \par Severe, Normal \par Severe, Normal \par Dry, Wet, Snow-covered & Categorical \par Categorical \par Categorical \par Categorical \\ \hline
Driving Behavior          & Lane Keeping Quality \par \vspace{0.2cm} Time-to-collision Riskiness & Good, Bad, Worst \par \vspace{0.2cm} \(\frac{1}{TTC} \cdot \max(0, \frac{2.2 - |Lateral\space Position|}{2.2})\) \vspace{0.2cm} & Categorical \par \vspace{0.2cm} Continuous \\ \hline
\end{tabular}
\end{table*}

Our dataset comprises of \textit{vehicle bus signals} and \textit{perception outputs}, as described in Section~\ref{subsec:workflow}. After processing and feature engineering, the final dataset, referred to as the Driving Anomaly Data (DAD), is used as input for the anomaly detection model. DAD is a multivariate tabular dataset, where each row represents a single 6-second window, and each column corresponds to a specific feature. The rows are randomly sampled to avoid sequential biases, and the extracted features are detailed in Table~\ref{tab:features_characteristics}. These features were selected based on their ability to capture three key aspects: i) vehicle driving kinematics, ii) environmental factors, and iii) driving behavior. While there are numerous features related to driving, this study focuses on the most essential ones for unsupervised anomaly detection.

For vehicle driving kinematics, we use the \textit{relative speed range} as the primary feature, as it effectively reflects variations in speed over the window while accounting for proportional changes, making it more robust than \textit{speed range}. For example, braking from 200 km/h to 180 km/h is more common and less likely to be considered anomalous than braking from 30 km/h to 10 km/h. With \textit{speed range}, both cases have a value of 20 km/h, making it impossible to distinguish anomalies. However, with \textit{relative speed range}, the values are normalized to \(\frac{200-180}{200}=0.1\) and \(\frac{30-10}{30}=0.67\), respectively, thus making it easier to quantify how anomalous each case is.

For environmental factors, we use mode features derived from categorical signals, such as \textit{weather severity} and \textit{road-type conditions}. These categorical features are converted into binary values using one-hot encoding to integrate seamlessly with continuous features. Driving behavior is captured through \textit{lane-keeping quality}, aggregated from three lane boundary safety signals (left, middle, and right of ego's lane). The feature is categorized into three levels: good (all lane boundaries are safe), bad (one lane boundary is unsafe), and worst (all lane boundaries are unsafe). To estimate \textit{collision risk}, we derive a feature based on \textit{time-to-collision} (TTC) with detected vehicles and their \textit{lateral position} relative to the ego vehicle. A scenario is considered risky if the TTC is less than 2 seconds and the lateral position is less than 2.2 meters. The resulting collision risk feature ranges from 0 (not risky) to 1 (high risk), providing a continuous measure of risk level.

Given the dataset's large size and unlabeled nature, manual annotation is impractical. Instead, we create a rule-based proxy anomaly label set to serve as a heuristic anomaly reference set or unsupervised evaluation. We use the term "proxy" to refer to automatically generated labels based on heuristic rules. These do not represent manual annotations or verified ground truth. Using pre-defined heuristic rules, we filter the dataset to extract a subset representing proxy anomalies. Examples of these proxy rules include:

\begin{itemize}
    \item \textbf{Extreme speed variations}: extremely high relative speed range.
    
    \item \textbf{Unusual signal combinations}: severe rain on dry road, severe rain with severe sun ray, severe sun ray with blur image.
    
    \item \textbf{Risky events}: bad lane keeping under severe rain, bad lane keeping with high relative speed range, high relative speed range on wet or snow-covered road, high collision riskiness.
\end{itemize}

These rules are manually defined based on the extracted features and do not capture all possible anomalies in the dataset. However, they provide an effective way to annotate and filter data for evaluation or to inject synthetic anomalies. The size of the proxy set can be adjusted by modifying the thresholds defined for each signal in heuristic rules. These thresholds determine the criteria for selecting proxy anomalies. For instance, time-to-collision values below a critical threshold, lateral acceleration exceeding an upper threshold, or significant change in speed during adverse weather are considered anomalous. By modifying these signal-specific thresholds in the heuristic rules, we can adjust the number of proxy anomalies, with stricter thresholds resulting in fewer anomalies and relaxed thresholds capturing more.

%For instance, higher thresholds result in a proxy set containing only extremely rare events (e.g., 1\%), reducing contamination levels. Conversely, lowering the thresholds increases the number of proxy anomalies, offering flexibility in dataset creation for different contamination levels.

\subsection{Implementation Details}
\label{sec:implementation_details}

This section introduces the details of our approach's implementation. DIF uses 50 representations (\(r=50\)) and six isolation trees per representation (\(t=6\)), with a subsampling size of 256 (\(n=256\)) for each isolation tree. 

DIF processes tabular data using a fully connected multi-layer perceptron (MLP) network with two hidden layers of 500 and 100 units, respectively, and a \textit{tanh} activation function. The network outputs representations with 20 dimensions, with optional skip connections and dropout applied based on configuration. Weights in the network are initialized with a normal distribution (mean 0, standard deviation 1), and the network processes data in batches of 64 samples. Competing IF used for comparison in the experiments employ 300 trees (matching the ensemble size of DIF, \(50 \times 6\)) with the same subsampling size of 256. The training was conducted on an NVIDIA A100 GPU using CUDA for acceleration. All anomaly detection algorithms, including DIF, were implemented in Python, with IF from the scikit-learn package used as a baseline. The unsupervised nature of the method allowed it to take the entire unlabeled dataset as input for training, producing predictions for the same dataset. Additionally, the contamination level (the expected proportion of anomalies) was manually set to guide the model.

\begin{figure}[htbp]
    \centering
    \includegraphics[width=\linewidth]{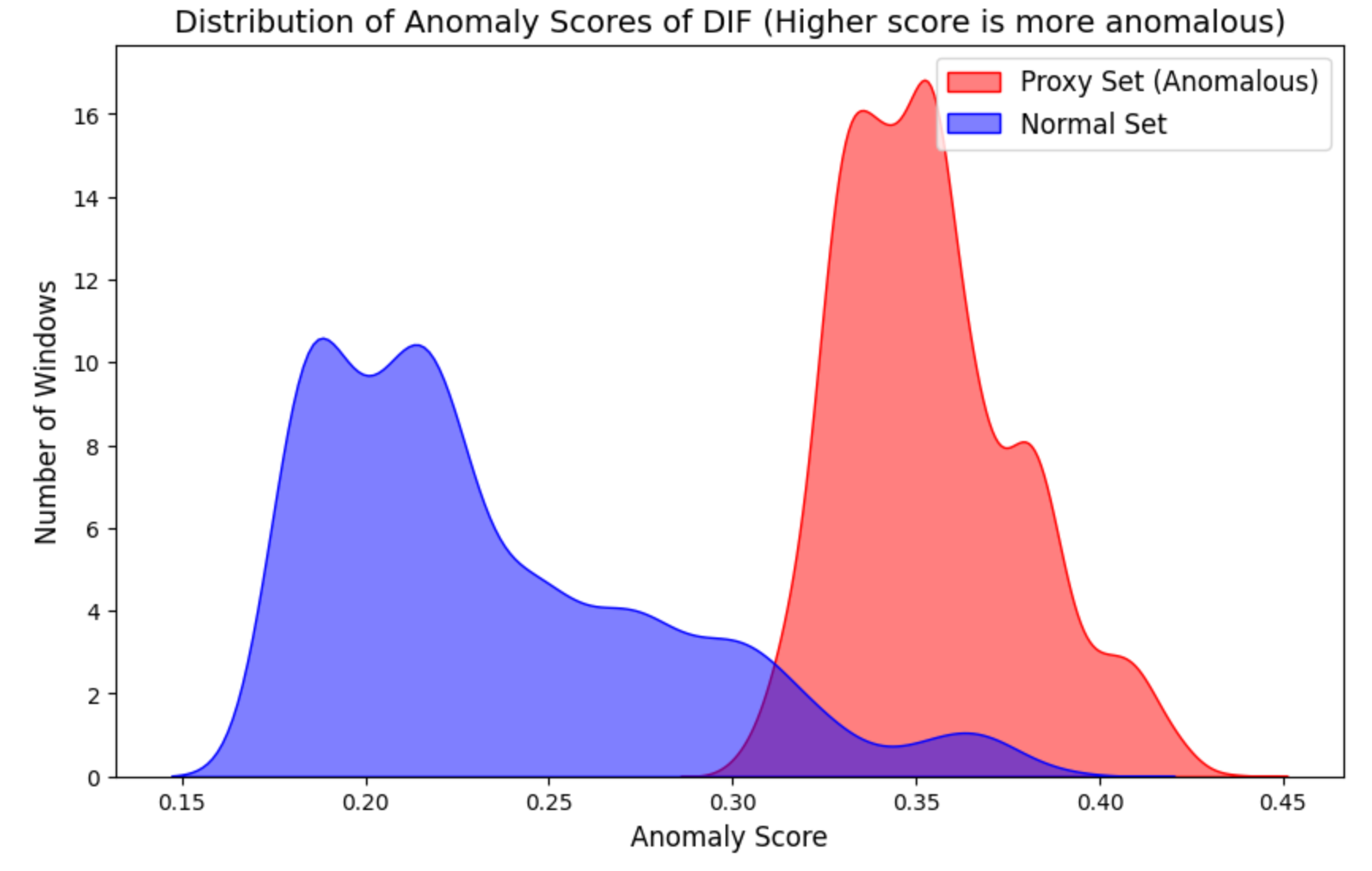} % Replace with the path to your image
    \caption{Anomaly score distributions for events labeled as anomalous by proxy heuristics vs. randomly sampled normal events. Higher scores indicate stronger anomaly signals by DIF}
    \label{fig:anomaly_score_distribution} % Replace with your desired label
\end{figure}

\begin{figure*}[htbp]
    \centering
    \includegraphics[width=\textwidth]{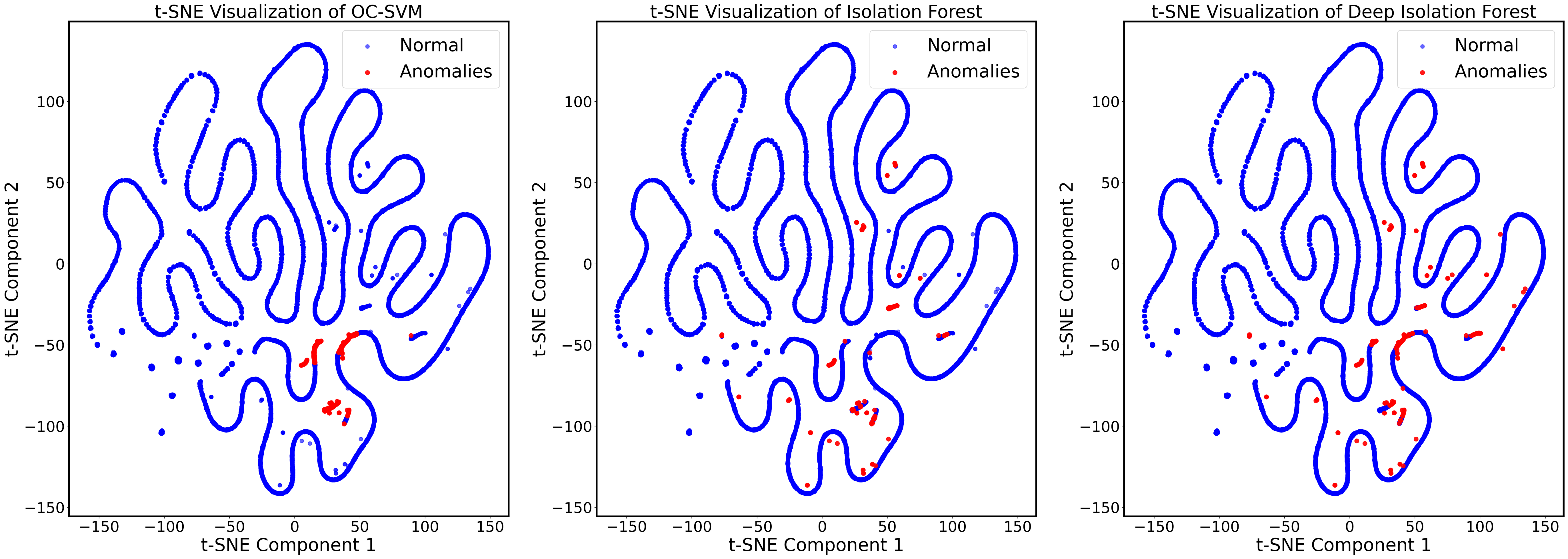} % Replace with the path to your image
    \caption{t-SNE visualization of OC-SVM (left), Isolation Forest (middle), and Deep Isolation Forest (right). Red points represent detected anomalies, while blue points indicate normal data. The figure demonstrates that OC-SVM performs the worst in capturing anomalies, followed by Isolation Forest. Deep Isolation Forest outperforms both by capturing significantly more outliers, highlighting its superior anomaly detection capability.}
    \label{fig:t_sne} % Replace with your desired label
\end{figure*}

\subsection{Evaluation}
\label{sec:evaluation}

Evaluating unsupervised algorithms is inherently more challenging than supervised methods because the absence of labeled data makes it difficult to measure performance directly. In this study, we use a proxy set (described in section~\ref{sec:dataset}) as ground truth anomaly annotations for the evaluation. We propose four evaluation approaches. The first analysis compares the distribution of the anomaly scores for events in the proxy set and normal sets. The second analysis examines the top 100 anomalous events based on their anomaly scores and evaluates how many of them overlap with events in the proxy set. The third analysis provides dimensionality reduction and visualization of the dataset to inspect anomalous data points interactively. Finally, we export some video data frames of the events with the highest anomaly scores to inspect whether they are meaningful anomalous scenarios perceptually.

\subsubsection{Anomaly Scores Distribution}
\label{sec:anomaly_scores_distribution}

The distribution of anomaly scores for the proxy set is compared against a normal set (randomly sampled) of equal size. In particular, we compare the anomaly score distributions between two sets: events in the proxy anomaly label set (expected to be anomalous) and events randomly sampled from the remaining dataset (expected to be normal). In our experiment, we identified 550 events using heuristic rules to define the proxy ground truth, representing approximately 2\% of the dataset. We then randomly sampled 550 additional events from the presumed normal set to compare their anomaly score distributions. Fig.~\ref{fig:anomaly_score_distribution} illustrates the anomaly score distributions for both sets. It shows that events in the proxy anomaly label set tend to have higher anomaly scores than those in the normal set. This higher anomaly score distribution for the proxy set indicates that the model effectively distinguishes anomalous events from normal ones.

\subsubsection{Top Anomalies Analysis}
\label{sec:top_anomalies_analysis}

The top 100 scenarios (i.e., events) with the highest anomaly scores are extracted. These scenarios are classified into two sets: those overlapping with the proxy anomaly label set (anomalous) and those outside it (normal). A summary table, as in Table~\ref{tab:anomaly_score_distribution}, shows the proportion of these high-scoring anomalies aligning with the proxy labels. The majority of the scenarios in the Top 100 group overlap with the proxy ground truth, while only a small percentage are classified as normal. In contrast, the Random 100 group contains a much higher percentage of scenarios from the normal set. This shows that our model effectively identifies most of the anomalies in the proxy set.

\begin{table}[h]
\renewcommand{\arraystretch}{1.3}
\caption{Assignment of the top 100 highest-scoring anomaly events detected by DIF and IF into the normal and proxy sets.}
\label{tab:anomaly_score_distribution}
\centering
\begin{tabular}{|c|c|c|}
\hline
\textbf{Events} & \textbf{Normal set} & \textbf{Proxy set} \\ \hline
Top 100 of DIF      & 16      & 84      \\ \hline
Top 100 of IF     & 35      & 65      \\ \hline
Random 100      & 88      & 12      \\ \hline
\end{tabular}
\end{table}

\subsubsection{t-SNE}
\label{sec:t_sne_evaluation}
To further evaluate the separation between anomalous and normal scenarios, we applied t-SNE to project the high-dimensional feature space into a two-dimensional visualization. Fig.~\ref{fig:t_sne} presents the t-SNE visualizations of the dataset for OC-SVM \cite{10.1145/2500853.2500857} (left), Isolation Forest (middle), and Deep Isolation Forest (right). Red points represent detected anomalous scenarios, while blue points indicate normal scenarios. The results illustrate that OC-SVM performs poorly; Isolation Forest struggles with some challenging local outliers, whereas Deep Isolation Forest outperforms both by effectively detecting these anomalies.

\subsubsection{Perceptual Evaluation}
\label{sec:perceptual_evaluation}

\begin{figure*}[htbp]
    \centering
    \begin{minipage}[b]{0.32\textwidth}
        \centering
        \includegraphics[width=\textwidth]{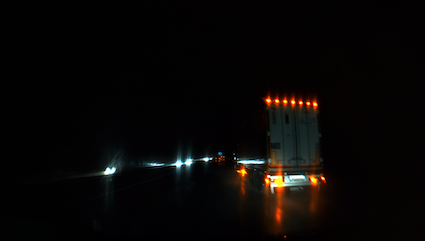}
        \caption{High speed on slippery roads with unsafe lane boundaries}
        \label{fig:anomaly_example_1}
    \end{minipage}
    \hfill
    \begin{minipage}[b]{0.32\textwidth}
        \centering
        \includegraphics[width=\textwidth]{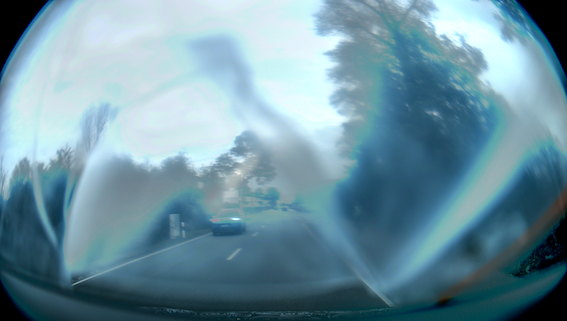}
        \caption{Low time-to-collision under blurry image conditions}
        \label{fig:anomaly_example_2}
    \end{minipage}
    \hfill
    \begin{minipage}[b]{0.32\textwidth}
        \centering
        \includegraphics[width=\textwidth]{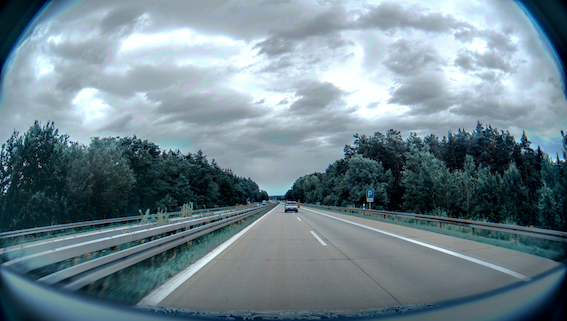}
        \caption{Sudden braking triggered by the front vehicle's abrupt lane change without a turn signal}
        \label{fig:anomaly_example_3}
    \end{minipage}
    \caption{Exported video frames from anomalous scenarios}
    \label{fig:anomaly_examples}
\end{figure*}

Lastly, frames corresponding to extreme anomalies are exported for human inspection. This step provides a qualitative assessment of the model's performance, allowing domain experts to verify whether the identified anomalies represent genuinely interesting or rare scenarios. Given the time and resource-intensive nature of perceptual evaluations, we concentrated only on the most anomalous segments identified by the model. These segments were exported as image frames from video data and manually reviewed to determine whether they represent real anomalies, risky situations, or dangerous scenarios. The results are displayed in Fig.~\ref{fig:anomaly_examples}, including sub-figures representing interesting anomalous events. Fig.~\ref{fig:anomaly_example_1} depicts high speed on slippery roads with unsafe lane boundaries. Fig.~\ref{fig:anomaly_example_2} shows a scenario with low time-to-collision under blurry image conditions. Fig.~\ref{fig:anomaly_example_3} captures sudden braking triggered by a front vehicle's abrupt, unsignaled lane change, illustrating the hazards of unpredictable driving behavior. This perceptual evaluation confirms that the proposed unsupervised approach effectively identifies anomalous events. The high anomaly scores correspond to genuine hazards or rare scenarios, reinforcing the method's utility in detecting meaningful anomalies within the data.

%!TeX spellcheck = en_US
% !TeX root = ../main.tex
\section{Discussion}
\label{sec:discussion}

The experimental results demonstrate that Deep Isolation Forest (DIF) outperforms baselines in IF and classical SVMs regarding the anomaly detection. We could display, that DIF captures more local outliers than IF, and the Top 100 analysis indicates that DIF detects 84\% of proxy anomalies, compared to 65\% by IF. This improvement is because DIF integrates deep neural networks to transform the data into random representation spaces, enabling it to detect hard anomalies that can only be easily isolated in higher-order subspaces. While our proposed framework demonstrates strong performance in detecting rare driving scenarios, several limitations remain. 

First, the reliance on a heuristic-based proxy ground truth means that evaluation is inherently limited by the quality and coverage of those rules. These heuristics may miss subtle or complex anomalies and introduce biases based on feature selection. Therefore, refining the proxy ground truth with more sophisticated rules to capture complex anomalies is crucial for better evaluation. Approaches like rule-based systems augmented with machine learning methods for driver fingerprinting \cite{8863987} could enhance the accuracy and utility of the proxy ground truth.  

Second, although the Deep Isolation Forest model captures non-linear anomalies effectively, its performance is sensitive to hyperparameters such as window size, contamination level, and feature representation. A systematic sensitivity analysis was beyond the current scope but is critical for robust deployment.

Third, our current framework uses manually engineered features, which may not fully capture the high-dimensional contextual relationships in driving behavior. Future work should explore automated feature learning methods, such as Deep Feature Synthesis \cite{7344858}  or representation learning, to reduce reliance on domain-specific signal selection. 

Finally, while t-SNE has proven useful for visualization, it is a non-deterministic and parameter-sensitive technique, which limits its interpretability and reproducibility. Future research should consider more stable and scalable alternatives like UMAP, and investigate automatic tuning strategies for visualization parameters.

\section{Conclusion}
\label{sec:conclusion}

This study presents an unsupervised framework for detecting rare and extreme driving scenarios in naturalistic driving data. By leveraging Deep Isolation Forest (DIF), we address key limitations of traditional rule-based and linear anomaly detection methods, enabling the identification of complex, non-linear anomalies. Our evaluation framework utilizes a proxy ground truth set, allowing effective assessment of the unsupervised model while minimizing the need for time-intensive annotations or synthetic data injection. Despite its strengths, challenges remain in refining feature selection, optimizing hyperparameters, and improving the quality and coverage of the proxy ground truth set. Future work will focus on enhancing feature engineering to capture richer contextual information, automating parameter optimization to improve model performance, and optimizing visualization parameters to interpret anomalous patterns better. Another research direction will be exploring the  impact of foundational models \cite{gao2025words,gao2025foundation}.

\bibliographystyle{IEEEtran}
\bibliography{literatur}

@INPROCEEDINGS{5482295,
  author={Dai, Jiangpeng and Teng, Jin and Bai, Xiaole and Shen, Zhaohui and Xuan, Dong},
  booktitle={2010 4th International Conference on Pervasive Computing Technologies for Healthcare}, 
  title={Mobile phone based drunk driving detection}, 
  year={2010},
  volume={},
  number={},
  pages={1-8},
  doi={10.4108/ICST.PERVASIVEHEALTH2010.8901}}

@INPROCEEDINGS{6232298,
  author={Eren, H. and Makinist, S. and Akin, E. and Yilmaz, A.},
  booktitle={2012 IEEE Intelligent Vehicles Symposium}, 
  title={Estimating driving behavior by a smartphone}, 
  year={2012},
  volume={},
  number={},
  pages={234-239},
  doi={10.1109/IVS.2012.6232298}}

@INPROCEEDINGS{7338354,
  author={Chen, Zhongyang and Yu, Jiadi and Zhu, Yanmin and Chen, Yingying and Li, Minglu},
  booktitle={2015 12th Annual IEEE International Conference on Sensing, Communication, and Networking (SECON)}, 
  title={D3: Abnormal driving behaviors detection and identification using smartphone sensors}, 
  year={2015},
  volume={},
  number={},
  pages={524-532},
  doi={10.1109/SAHCN.2015.7338354}}

@ARTICLE{7116534,
  author={Wahlström, Johan and Skog, Isaac and Händel, Peter},
  journal={IEEE Transactions on Intelligent Transportation Systems}, 
  title={Detection of Dangerous Cornering in GNSS-Data-Driven Insurance Telematics}, 
  year={2015},
  volume={16},
  number={6},
  pages={3073-3083},
  doi={10.1109/TITS.2015.2431293}}

@INPROCEEDINGS{7795864,
  author={Fu Li and Hai Zhang and Huan Che and Xiaochen Qiu},
  booktitle={2016 IEEE 19th International Conference on Intelligent Transportation Systems (ITSC)}, 
  title={Dangerous driving behavior detection using smartphone sensors}, 
  year={2016},
  volume={},
  number={},
  pages={1902-1907},
  doi={10.1109/ITSC.2016.7795864}}

@inproceedings{inproceedings1,
author = {Mohan, Prashanth and Padmanabhan, Venkata and Ramjee, Ramachandran},
year = {2008},
month = {11},
pages = {323-336},
title = {Nericell: Rich monitoring of road and traffic conditions using mobile smartphones},
journal = {Proceedings of the 6th ACM Conference on Embedded Network Sensor Systems},
doi = {10.1145/1460412.1460444}
}

@INPROCEEDINGS{9095790,
  author={Yang, Chule and Renzaglia, Alessandro and Paigwar, Anshul and Laugier, Christian and Wang, Danwei},
  booktitle={2019 IEEE International Conference on Cybernetics and Intelligent Systems (CIS) and IEEE Conference on Robotics, Automation and Mechatronics (RAM)}, 
  title={Driving Behavior Assessment and Anomaly Detection for Intelligent Vehicles}, 
  year={2019},
  volume={},
  number={},
  pages={524-529},
  doi={10.1109/CIS-RAM47153.2019.9095790}}

@article{article1,
author = {Liu, Zhenyu and Wu, Mengfei and Zhu, Konglin and Zhang, Lin},
year = {2016},
month = {10},
pages = {1-13},
title = {SenSafe: A Smartphone-Based Traffic Safety Framework by Sensing Vehicle and Pedestrian Behaviors},
volume = {2016},
journal = {Mobile Information Systems},
doi = {10.1155/2016/7967249}
}

@INPROCEEDINGS{9304789,
  author={Breitenstein, Jasmin and Termöhlen, Jan-Aike and Lipinski, Daniel and Fingscheidt, Tim},
  booktitle={2020 IEEE Intelligent Vehicles Symposium (IV)}, 
  title={Systematization of Corner Cases for Visual Perception in Automated Driving}, 
  year={2020},
  volume={},
  number={},
  pages={1257-1264},
  doi={10.1109/IV47402.2020.9304789}}

@inproceedings{inproceedings2,
author = {Chakravarty, Tapas and Ghose, Avik and Bhaumik, Chirabrata and Chowdhury, Arijit},
year = {2013},
month = {12},
pages = {338-344},
title = {MobiDriveScore — A system for mobile sensor based driving analysis: A risk assessment model for improving one's driving},
isbn = {978-1-4673-5222-2},
doi = {10.1109/ICSensT.2013.6727671}
}

@ARTICLE{10108034,
  author={Xu, Hongzuo and Pang, Guansong and Wang, Yijie and Wang, Yongjun},
  journal={IEEE Transactions on Knowledge and Data Engineering}, 
  title={Deep Isolation Forest for Anomaly Detection}, 
  year={2023},
  volume={35},
  number={12},
  pages={12591-12604},
  doi={10.1109/TKDE.2023.3270293}}

@article{JMLR:v9:vandermaaten08a,
  author  = {Laurens van der Maaten and Geoffrey Hinton},
  title   = {Visualizing Data using t-SNE},
  journal = {Journal of Machine Learning Research},
  year    = {2008},
  volume  = {9},
  number  = {86},
  pages   = {2579--2605},
  url     = {http://jmlr.org/papers/v9/vandermaaten08a.html}
}

@article{DBLP:journals/corr/abs-2103-03678,
  author       = {Florian Heidecker and
                  Jasmin Breitenstein and
                  Kevin R{\"{o}}sch and
                  Jonas L{\"{o}}hdefink and
                  Maarten Bieshaar and
                  Christoph Stiller and
                  Tim Fingscheidt and
                  Bernhard Sick},
  title        = {An Application-Driven Conceptualization of Corner Cases for Perception
                  in Highly Automated Driving},
  journal      = {CoRR},
  volume       = {abs/2103.03678},
  year         = {2021},
  url          = {https://arxiv.org/abs/2103.03678},
  eprinttype    = {arXiv},
  eprint       = {2103.03678},
  bibsource    = {dblp computer science bibliography, https://dblp.org}
}

@ARTICLE{4815488,
  author={Malta, Lucas and Miyajima, Chiyomi and Takeda, Kazuya},
  journal={IEEE Transactions on Intelligent Transportation Systems}, 
  title={A Study of Driver Behavior Under Potential Threats in Vehicle Traffic}, 
  year={2009},
  volume={10},
  number={2},
  pages={201-210},
  doi={10.1109/TITS.2009.2018321}}

@INPROCEEDINGS{6831046,
  author={Hongyang Zhao and Huan Zhou and Canfeng Chen and Chen, Jiming},
  booktitle={2013 IEEE Global Communications Conference (GLOBECOM)}, 
  title={Join driving: A smart phone-based driving behavior evaluation system}, 
  year={2013},
  volume={},
  number={},
  pages={48-53},
  doi={10.1109/GLOCOM.2013.6831046}}

@inproceedings{inproceedings3,
author = {Hofmockel, Julia and Sax, Eric},
year = {2018},
month = {01},
pages = {411-416},
title = {Isolation Forest for Anomaly Detection in Raw Vehicle Sensor Data},
doi = {10.5220/0006758004110416}
}

@INPROCEEDINGS{8417777,
  author={Matousek, Matthias and Yassin, Mahmoud and Al-Momani, Ala'a and van der Heijden, Rens and Kargl, Frank},
  booktitle={2018 IEEE 87th Vehicular Technology Conference (VTC Spring)}, 
  title={Robust Detection of Anomalous Driving Behavior}, 
  year={2018},
  volume={},
  number={},
  pages={1-5},
  doi={10.1109/VTCSpring.2018.8417777}}

@INPROCEEDINGS{7795771,
  author={Zheng, Yang and Hansen, John H.L.},
  booktitle={2016 IEEE 19th International Conference on Intelligent Transportation Systems (ITSC)}, 
  title={Unsupervised driving performance assessment using free-positioned smartphones in vehicles}, 
  year={2016},
  volume={},
  number={},
  pages={1598-1603},
  doi={10.1109/ITSC.2016.7795771}}

@INPROCEEDINGS{4781136,
  author={Liu, Fei Tony and Ting, Kai Ming and Zhou, Zhi-Hua},
  booktitle={2008 Eighth IEEE International Conference on Data Mining}, 
  title={Isolation Forest}, 
  year={2008},
  volume={},
  number={},
  pages={413-422},
  doi={10.1109/ICDM.2008.17}}

@misc{bouman2023unsupervisedanomalydetectionalgorithms,
      title={Unsupervised anomaly detection algorithms on real-world data: how many do we need?}, 
      author={Roel Bouman and Zaharah Bukhsh and Tom Heskes},
      year={2023},
      eprint={2305.00735},
      archivePrefix={arXiv},
      primaryClass={cs.LG},
      url={https://arxiv.org/abs/2305.00735}, 
}

@article{doi:10.1080/14786440109462720,
author = { Karl   Pearson   F.R.S. },
title = {LIII. On lines and planes of closest fit to systems of points in space},
journal = {The London, Edinburgh, and Dublin Philosophical Magazine and Journal of Science},
volume = {2},
number = {11},
pages = {559-572},
year  = {1901},
publisher = {Taylor & Francis},
doi = {10.1080/14786440109462720},
}

@INPROCEEDINGS{7344858,
  author={Kanter, James Max and Veeramachaneni, Kalyan},
  booktitle={2015 IEEE International Conference on Data Science and Advanced Analytics (DSAA)}, 
  title={Deep feature synthesis: Towards automating data science endeavors}, 
  year={2015},
  volume={},
  number={},
  pages={1-10},
  doi={10.1109/DSAA.2015.7344858}}

@ARTICLE{8863987,
  author={Xun, Yijie and Liu, Jiajia and Kato, Nei and Fang, Yongqiang and Zhang, Yanning},
  journal={IEEE Transactions on Industrial Informatics}, 
  title={Automobile Driver Fingerprinting: A New Machine Learning Based Authentication Scheme}, 
  year={2020},
  volume={16},
  number={2},
  pages={1417-1426},
  doi={10.1109/TII.2019.2946626}}

@inproceedings{Liu2010OnDC,
  title={On Detecting Clustered Anomalies Using SCiForest},
  author={Fei Tony Liu and Kai Ming Ting and Zhi-Hua Zhou},
  booktitle={ECML/PKDD},
  year={2010},
  url={https://api.semanticscholar.org/CorpusID:5721991}
}

@article{Hariri_2021,
   title={Extended Isolation Forest},
   volume={33},
   ISSN={2326-3865},
   number={4},
   journal={IEEE Transactions on Knowledge and Data Engineering},
   publisher={Institute of Electrical and Electronics Engineers (IEEE)},
   author={Hariri, Sahand and Kind, Matias Carrasco and Brunner, Robert J.},
   year={2021},
   month=apr, pages={1479–1489} }

@article{article5,
author = {Lesouple, Julien and Baudoin, Cédric and Spigai, M. and Tourneret, Jean-Yves},
year = {2021},
month = {06},
pages = {},
title = {Generalized Isolation Forest for Anomaly Detection},
volume = {149},
journal = {Pattern Recognition Letters},
doi = {10.1016/j.patrec.2021.05.022}
}

@article{10.1016/j.ins.2021.10.075,
author = {Tokovarov, Mikhail and Karczmarek, Pawe\l{}},
title = {A probabilistic generalization of isolation forest},
year = {2022},
issue_date = {Jan 2022},
publisher = {Elsevier Science Inc.},
address = {USA},
volume = {584},
number = {C},
issn = {0020-0255},
url = {https://doi.org/10.1016/j.ins.2021.10.075},
doi = {10.1016/j.ins.2021.10.075},
journal = {Inf. Sci.},
month = jan,
pages = {433–449},
numpages = {17}
}

@inproceedings{10.1145/2500853.2500857,
author = {Amer, Mennatallah and Goldstein, Markus and Abdennadher, Slim},
title = {Enhancing one-class support vector machines for unsupervised anomaly detection},
year = {2013},
isbn = {9781450323352},
publisher = {Association for Computing Machinery},
address = {New York, NY, USA},
url = {https://doi.org/10.1145/2500853.2500857},
doi = {10.1145/2500853.2500857},
booktitle = {Proceedings of the ACM SIGKDD Workshop on Outlier Detection and Description},
pages = {8–15},
numpages = {8},
location = {Chicago, Illinois},
series = {ODD '13}
}

@article{gao2025foundation,
  title={Foundation Models in Autonomous Driving: A Survey on Scenario Generation and Scenario Analysis},
  author={Gao, Yuan and Piccinini, Mattia and Zhang, Yuchen and Wang, Dingrui and Moller, Korbinian and Brusnicki, Roberto and Zarrouki, Baha and Gambi, Alessio and Totz, Jan Frederik and Storms, Kai and others},
  journal={arXiv preprint arXiv:2506.11526},
  year={2025}
}

@article{gao2025words,
  title={From Words to Collisions: LLM-Guided Evaluation and Adversarial Generation of Safety-Critical Driving Scenarios},
  author={Gao, Yuan and Piccinini, Mattia and Moller, Korbinian and Alanwar, Amr and Betz, Johannes},
  journal={arXiv preprint arXiv:2502.02145},
  year={2025}
}

@inproceedings{wang2024esp,
  title={Esp: Extro-spective prediction for long-term behavior reasoning in emergency scenarios},
  author={Wang, Dingrui and Lai, Zheyuan and Li, Yuda and Wu, Yi and Ma, Yuexin and Betz, Johannes and Yang, Ruigang and Li, Wei},
  booktitle={2024 IEEE International Conference on Robotics and Automation (ICRA)},
  pages={13030--13037},
  year={2024},
  organization={IEEE}
}
\end{document}